%% file: acl2021.tex
\documentclass[11pt,a4paper]{article}
\usepackage[hyperref]{acl2021}
\usepackage{times}
\usepackage{latexsym}

\usepackage{microtype}
\usepackage{booktabs}
\usepackage{tabularx}
\usepackage{multirow}
\usepackage{graphicx}
\usepackage{footnote}
\usepackage{amsmath}
\aclfinalcopy 

\newcommand\fb{$^\heartsuit$}
\newcommand\unc{$^\spadesuit$}
\newcommand\msr{$^\clubsuit$}

\title{\textsc{EmailSum}: Abstractive Email Thread Summarization}

\author{Shiyue Zhang\unc $\;\;\;\;$ Asli Celikyilmaz\fb $\;\;\;\;$ Jianfeng Gao\msr  $\;\;\;\;$ Mohit Bansal\unc \\
  \unc UNC Chapel Hill $\;\;$ \fb Facebook AI Research $\;\;$ \msr Microsoft Research \\
  {\tt \{shiyue, mbansal\}@cs.unc.edu} \\
  {\tt aslic@fb.com} $\;\;$ {\tt jfgao@microsoft.com} \\
}

\date{}

\begin{document}
\maketitle
\begin{abstract}
Recent years have brought about an interest in the challenging task of summarizing conversation threads (meetings, online discussions, etc.). Such summaries help analysis of the long text to quickly catch up with the decisions made and thus improve our work or communication efficiency. 
To spur research in thread summarization, we have developed an abstractive \textbf{Email} Thread \textbf{Sum}marization (\textsc{EmailSum}) dataset, which contains human-annotated short ($<$30 words) and long ($<$100 words) summaries of 2,549 email threads (each containing 3 to 10 emails) over a wide variety of topics.
We perform a comprehensive empirical study to explore different summarization techniques (including extractive and abstractive methods, single-document and hierarchical models, as well as transfer and semi-supervised learning) and conduct human evaluations on both short and long summary generation tasks.
Our results reveal the key challenges of current abstractive summarization models in this task, such as understanding the sender's intent and identifying the roles of sender and receiver. 
Furthermore, we find that widely used automatic evaluation metrics (ROUGE, BERTScore) are weakly 
correlated with human judgments on this email thread summarization task. Hence, we emphasize the importance of human evaluation and the development of better metrics by the community.\footnote{Our code and summary data have been made available at: \url{https://github.com/ZhangShiyue/EmailSum}}
\end{abstract}

\input{introduction}

\input{data}
\input{models}
\input{experiments}

\input{human_eval}

\section{Conclusion}
In this work, we propose an abstractive email thread summarization dataset, \textsc{EmailSum}, that contains 2,549 email threads with human-written short and long summaries. We explore different summarization paradigms and find that taking the email thread as a single document and finetuning T5  \cite{raffel2020exploring} sets up a good baseline. Transferring from other summarization datasets barely improves it. Using hierarchical structure also only marginally improves the performance. Semi-supervised learning by using unlabelled email threads improves automatic metrics (ROUGE) but still loses to the baseline in human evaluation. Finally, our human evaluation reveals that the model fails to understand the sender's main intention and the roles of different speakers. Automatic metrics are poorly correlated with human judgment, which emphasizes the importance of human evaluation and designing new metrics for this task in the future.

\section{Broader Impact Statement}
We use two email collections in this work: Avocado \cite{linguistic2015avocado} and W3C \cite{craswell2006overview}. W3C is derived from W3C Public Mailing List that is open-source available online. Avocado consists of emails and attachments taken from 279 accounts of a defunct information technology company referred to as ``Avocado''. Its copyright is protected by Linguistic Data Consortium. 
Based on the license agreement, we will only open-source our collected summaries and provide scripts to obtain email threads from the original Avocado email collection. 
To further protect copyright and the privacy of the persons involved in the emails, as introduced in Section~\ref{data}, we carefully anonymize all the email threads we construct from both email collections. We fairly pay crowd-source workers
\$1.37 (for threads with 5 or fewer emails) or \$2 (for threads with more than 5 emails) for writing the short and long summaries and \$0.6 for human rating such that the pay rate is higher than the federal minimum wage requirement. 

\section*{Acknowledgments}
We thank the reviewers for their helpful comments and Xiang Zhou for useful discussions. We thank Saadia Gabriel, Yichen Jiang, Tom McCoy, and Yixin Nie for helping write summary examples (to show as initial examples to MTurk annotators) and estimate the workload for deciding the fair payment. This work was partially done while SZ was interning at MSR and later extended at UNC, where it was supported by NSF-CAREER Award 1846185, ONR Grant N00014-18-1-2871, and a Microsoft Investigator Fellowship. The views contained in this article are those of the authors and not of the funding agency.

\bibliographystyle{acl_natbib}
\bibliography{acl2021}

\input{appendix}

\end{document}

%% file: introduction.tex
\section{Introduction}
As one of the major natural language generation tasks, automatic summarization has been studied for decades. Most research efforts were focused on single-document summarization tasks, e.g., news document summarization \cite{hermann2015teaching, narayan2018don}.
However, living in an information era, we are facing with diverse content in different structures. The summarization need is varied along with different application scenarios. Recently, there is an increasing research interest in diverse summarization tasks \cite{ijcai2020-676}, e.g., timeline \cite{allan2001temporal}, query-based \cite{li2014query}, multi-modal \cite{zhu2018msmo}, meeting \cite{10.1007/11677482_3}, dialogue or discussion thread \cite{misra2015using, gliwa2019samsum, rameshkumar2020storytelling}, etc. Following the branch of dialogue or thread summarization, we introduce a new abstractive \textbf{Email} Thread \textbf{Sum}marization (\textsc{EmailSum}) dataset.

\begin{table}
\begin{center}
\small
\begin{tabular}{p{0.45\textwidth}}
\toprule 
\textbf{Email Thread}: \\
\emph{Subject}: lunch this week \\
\emph{Susan}: All, Regarding our lunch this week to celebrate the one year anniversaries for Michelle \& David, and Mark's birthday, I have a request to make it Wednesday instead of Tuesday.  Does anyone have an objection to this? Susan \\
\emph{David}: I have another lunch engagement Wed, but I will skip it if everyone else wants to move our lunch. David \\
\emph{Tamra}: Susan, Wednesday works out better for me as well.  I have a doctor's appointment tomorrow during lunch. Tamra\\
\midrule
\textbf{Short Summary}: \\
Susan emails everyone about an anniversary and offers to change the date. David says he is busy but is willing to go with the majority. Tamra agrees with Susan's date.\\
\midrule
\textbf{Long Summary}: \\
Susan emails everyone about a lunch to celebrate a one year anniversary as well as Mark's birthday. She says she would change the date to a different day. David says he is busy that day with his own appointment but is willing to go with the majority and cancel that appointment to make this one.  Tamra agrees with Susan's date as she is busy Tuesday with an appointment. \\
\bottomrule
\end{tabular}
\end{center}
\vspace{-5pt}
\caption{An email thread and human-written short and long summaries from our \textsc{EmailSum} Dataset. 
}
\vspace{-5pt}
\label{tab:lead-example}
\end{table}

Email threads are widely used at work. An email thread is a special type of dialogue that usually has a specific structure (sender, receiver, greeting line, main body, and the signature), contains technical information, and involves multiple speakers. Unlike a conversational dialog turn, an email in a thread is much longer with longer sentences, multiple action items or requests, and stylistically similar to written text. 
Studies have shown that on average a worker sends/receives 122 business emails \cite{Radicati2015} and spends more than 3 hours on those emails \cite{adobe2019} per day. One possible reason is that sometimes people have to read through the entire conversation before replying to the latest email. This happens when you forget the main points of previous discussions or you are newly included in a discussion thread. Therefore, automatically summarizing email threads can improve our work efficiency and provides practical benefits. Email Thread Summarization is not a new task. \citet{carenini2007summarizing} collected extractive summaries of 39 email threads from Enron email corpus \cite{klimt2004enron} and proposed to use a fragment quotation graph and clue words to conduct summarization. \citet{ulrich2008publicly} collected both extractive and abstractive summaries of 40 threads from W3C email corpus \cite{craswell2006overview} plus speech acts, meta sentences, etc. However, this task has been much less studied compared to other summarization tasks, partially due to the lack of large labeled email thread datasets. 

In this paper, we collect 
human-written short ($<$ 30 words) and long ($<$ 100 words) abstractive summaries of 2,549 email threads constructed from Avocado Research Email Collection \cite{linguistic2015avocado}, which is 64$\times$ the size of previously labeled email thread datasets \cite{carenini2007summarizing, craswell2006overview}. We limit each thread to a minimum of 3 and a maximum of 10 emails, an example is given in Table~\ref{tab:lead-example}. We also extract 8,594 unlabeled email threads from both Avocado and W3C to facilitate semi-supervised learning.\footnote{We apply strict criteria for thread extraction (see Section~\ref{data}). More threads can be extracted by relaxing those constraints.}  See Section~\ref{data} for details of data collection.

Next, we present comprehensive baselines from different learning paradigms as a benchmark for our new email summarization dataset. Specifically, we explore different summarization techniques, including extractive and abstractive summarization methods, single-document and hierarchical models, transfer learning, and semi-supervised learning for both short and long summary generation. Experiments demonstrate that utilizing pretrained language model (e.g., T5 \cite{raffel2020exploring}) is critical due to the small size of our data; taking the email thread as a single document sets up a good baseline; transferring from news or dialogue datasets barely improve the performance; using hierarchical encoders only marginally improves it; while semi-supervised learning by using unlabelled email threads significantly ($p<0.01$) improves ROUGE \cite{lin2004rouge} scores in some cases.

Lastly, to better understand how well the email thread summarization models perform and investigate the correlation between automatic metrics and human judgment, we ask humans to rate the ``salience'' (how well the model summarizes salient points) and ``faithfulness'' (how well the model stays true to the email thread) of model-generated summaries, as well as to perform a pairwise comparison between our best and base models. We find that even though semi-supervised learning improves ROUGE scores, human judges still favor the summary generated by the baseline model (T5$_{base}$). 
Two frequent errors made by the model are (1) \emph{failing to understand the sender's intent} and (2) \emph{failing to identify the roles of the sender and receiver}. Relatedly, 
human correlation analysis reveals that automatic metrics (ROUGE \cite{lin2004rouge}, BERTScore \cite{zhang2019bertscore}) are poorly correlated with human judgment, which stresses the importance of human evaluation in this task and the requirement for better metrics to be proposed. Overall, in this work, we propose the new \textsc{EmailSum} dataset that provides a larger resource for studying the email thread summarization task. We conduct a comprehensive empirical model study and human evaluation analysis, which will serve as an important starting point for future studies.

%% file: data.tex
\section{\textsc{EmailSum} Dataset}
\label{data}

To collect email thread summarization data, we first need to obtain unlabeled email threads. We resort to existing email collections:  Enron \cite{klimt2004enron}, W3C \cite{craswell2006overview}, and Avocado \cite{linguistic2015avocado}. However, none of them provides explicit thread structure. Therefore, in this section, we will introduce our email thread preprocessing and summary collection procedures. 

\begin{table*}
\begin{center}
\small
\begin{tabular}{llllllll}
\toprule 
Domain & \multicolumn{2}{c}{News} & \multicolumn{2}{c}{Dialogue} & \multicolumn{3}{c}{Email Thread} \\
 \cmidrule(lr){1-1} \cmidrule(lr){2-3} \cmidrule(lr){4-5} \cmidrule(lr){6-8}
Dataset & CNN/DM & XSum & SAMSum & CRD3 & BC3 & \textsc{EmailSum}$_{short}$ & \textsc{EmailSum}$_{long}$\\
 \midrule
 \# of documents & 312,085 & 226,677 & 16,369 & 32,720 & 40 & 2,549 & 2,549\\
 Avg. document length & 786.4 & 409.5 & 124.1 & 615.8 & 550.4 & 233.2 & 233.2\\
 \# of turns per doc. & - & - & 10.2 & 27.5 & 6.4 & 4.5 & 4.5 \\
 Avg. turn length & - & - & 11.1 & 19.4 & 85.3 & 50.3 & 50.3 \\
 Avg. summary length & 55.2 & 23.2 & 23.4 & 58.3 & 134.3 & 27.1 & 68.5\\
 Ext-Oracle-R1 & 58.2 & 23.8 & 45.3 & 50.4 & 36.5 & 39.0 & 46.0\\
\toprule
\end{tabular}
\end{center}
\vspace{-12pt}
\caption{The statistics of different summarization datasets. Ext-Oracle-R1s are the ROUGE-1 scores of the oracle extractive method, which shows the abstractiveness of the summary (the lower the more abstractive).}
\label{tab:stats}
\vspace{-12pt}
\end{table*}

\subsection{Email Thread Preprocessing}
\label{sec:data1}

We extract email threads from the flat email collections in the following steps: (1) we give every email a ``normalized subject'' by removing the reply or forward tags (e.g., ``Re:'', ``Fwd:'', etc.) from its original subject; (2) we group emails by the normalized subjects and sort emails in the same group (i.e., thread) by timestamp; (3) we de-duplicate emails in every thread by sender's email plus timestamp; (4) we traverse emails in every thread in temporal order and cut off the thread when none of the senders plus receivers of the current email appears in previous emails; (5) we filter out threads that only contain single repeated content.

To obtain a cleaner dataset, we remove threads that do not comply with the following constraints: (1) 3 $\leq$ the number of emails $\leq$ 10; (2) 5 $<$ the number of words in each email $<$ 200; (3) 30 $<$ the total number of words $<$ 1000;  
(4) does not contain non-English (e.g., German) tokens;
(5) does not contain reply or forward tags in the subject of the first email.

Emails often contain personal information such as full name, email/physical address, phone number, etc. To protect privacy, we anonymize all email threads before annotation: (1) only keep first names; (2) remove threads that have ``password'', ``pwd'', ``confidential'', etc.; (3) replace email address, physical address, phone number, URL, IP address, local path, and other sensitive numbers with USERNAME@DOMAIN.COM, ADDRESS, PHONENUMBER, HTTP://LINK, IPADDRESS, PATH, and NUMBER, respectively.

We conduct an extensive manual quality scan to make sure that the extracted threads are truly threads (instead of random emails grouped) and properly anonymized. Finally, we obtain 8,116 threads from Avocado and 3,478 threads from W3C.\footnote{We find that the extracted threads from Enron are usually short (fewer than 3 emails) and noisy.} We randomly sample 3K Avocado threads for summary annotation, and the remaining threads are used as unlabeled data.

\subsection{Thread Summary Collection}
We collect summary annotations on Amazon Mechanical Turk. Since summarizing text is not an easy task, to get acceptable English summaries we use several quality control strategies: (1) We select annotators that are located in the US, have an approval rate greater than 97\%, and have at least 10,000 approved HITs; (2) During annotation, we periodically sample summaries, manually check their quality, and reject or block poor-quality annotators; (3) After annotation, we randomly sample 2 examples per annotator and manually categorize annotators into ``good'', ``fair'', and ``bad'' groups, then filter examples written by bad annotators.

Email threads oftentimes contain technical information, we instruct annotators not to get stuck on technical details, instead, focus on the major concerns, decisions, and consensus. We collect both short ($< 30$ words) and long ($<$ 100 words) abstractive summaries per thread. 
For the short summary, we instruct annotators to write a \emph{concise description of what the thread is mainly talking about}; while for the long summary, we instruct them to write a \emph{narrative of what happens}. We are intent to provide summaries with two different levels of abstractiveness, length, and concreteness.  We show annotators an example written by an expert (a CS graduate student). More summary collection details can be found in Appendix~\ref{appedix:summay_annotation}.

\subsection{Final Dataset Description}
The summary collection and filtering process yield 2,549 email threads each with a long and a short summary. We randomly sample 500 examples from the ``good'' annotator group as our testing set and split the remaining examples into training (1,800 threads) and development (249 threads) sets. Table~\ref{tab:stats} shows the statistics of \textsc{EmailSum}.\footnote{Since comparing the model-generated summary to only one human-written reference may not be fully informative, recently we have also collected \emph{one more reference} for each email thread in our test set, i.e., each test example will have two gold references now in our final dataset. The results in the paper are all still based on the original one-reference setup but we will release the updated two-reference results for our best baselines on Github.}. For ease of benchmarking, we also include statistics on other commonly used summarization datasets: CNN/DM \cite{hermann2015teaching} and XSum \cite{narayan2018don} are about news summarization; SAMSum \cite{gliwa2019samsum} is about chit-chat summarization; CRD3 \cite{rameshkumar2020storytelling} is a role-play dialogue summarization dataset; BC3 \cite{ulrich2008publicly} is another email thread summarization with 40 threads from W3C. 
Compared to the other datasets, the average document length in the \textsc{EmailSum} dataset is not very long, containing 233 words; long summaries are more than twice as longer than short summaries. ``Ext-Oracle-R1'' in Table~\ref{tab:stats} indicates how abstractive the summaries are. It computes the ROUGE-1 scores of an oracle extractive method (see Section~\ref{extractive} for details of the oracle extractive method). The lower it is, the more abstractive the dataset is. According to this score, the abstractiveness of the \textsc{EmailSum} summaries is lower than the XSum summaries, while higher than the CNNDM summaries. Furthermore, the short summaries of \textsc{EmailSum} dataset are more abstractive than its long summaries.

%% file: models.tex
\section{Models}
The summarization models we explore in this work take the email thread as input and generate the summary as output.
We experiment on \textsc{EmailSum}$_{short}$ and \textsc{EmailSum}$_{long}$ tasks separately.

\subsection{Extractive}
\label{extractive}
\paragraph{Oracle.} 
This method 
maximize an evaluation metric w.r.t. the gold summary. ``Ext-Oracle-R1'' in Table~\ref{tab:stats}
is computed from an oracle summary that maximizes ROUGE-1 \cite{lin2004rouge}.

\paragraph{Lead.} This model simply picks the first sentence from the source document as the summary, which has surprisingly good performance on CNN/DM dataset \cite{narayan2018don}. We test two variants by selecting: (1) the first sentence of the email thread, which is usually the subject (see the example in Table~\ref{tab:lead-example}), referred as \textbf{Lead-1}; (2)  the first sentence of the email thread (the subject) plus the first sentences of every email, named \textbf{Lead-1-Email}.\footnote{We also tested some other heuristics: e.g., the first sentence of the last email, the last 3-5 sentences of the email thread, etc. However, none of them perform better than Lead-1-Email.}

\paragraph{TextRank.} This is a graph-based method \cite{mihalcea2004textrank}. It first builds a graph between sentences by their embedding similarities; then the PageRank algorithm is applied to obtain the rank scores for each sentence, and top-rank sentences are selected as the summary.

\paragraph{BertSumExt.} \citet{liu2019text} propose to build a sentence extractor upon BERT \cite{devlin-etal-2019-bert} to perform extractive summarization, which achieves a good performance on CNN/DM.

\subsection{Abstractive}

\paragraph{Fast Abs RL.} 
As the simple non-pretrained abstractive baseline, we use \citet{chen2018fast}, which is a hybrid model that first extracts sentences from the source document, then rewrites the extracted sentences by an abstractive rewriter. 
They pair summary sentences with the extracted sentences to train the abstractive rewriter. Adapting their model to our email thread summarization task, we make two adjustments: (1) We extract emails instead of sentences, which is a natural unit for email thread; (2) Since summary sentences usually follow the temporal order of the emails, we enhance this pairing procedure by using the Neeleman-Wunsch algorithm \cite{needleman1970general, rameshkumar2020storytelling} to impose the order constraint to the alignment (see description and comparison in Appendix~\ref{appedix:fastabs}).

\paragraph{T5.} 
T5 \cite{raffel2020exploring} is a Transformer \cite{vaswani2017attention} based seq-to-seq model pretrained with large-scale English data. It achieves state-of-the-art performances on a lot of NLP tasks including the CNN/DM summarization task. As our main baseline, we take the email thread as a single document and finetune a T5 base to generate the summary (\textbf{T5$_{base}$}). A similar setup is also used in transfer and semi-supervised learning. Since our training dataset is small, we find that using the pre-trained knowledge transfer is crucial. 
Training a T5 model from scratch performs poorly (see the results in Appendix Table~\ref{tab:dev}).  

\paragraph{Transfer Learning.}
To analyze how information from other summarization datasets (listed in Table~\ref{tab:stats}) can be transferred to this new task and its impact on the performance, we investigate two simple transfer learning methods: (1) \emph{Pre-finetuning}, in which we first finetune T5 on a bigger summarization dataset (e.g., CNN/DM) then continue the finetuning on our dataset, referred as \textbf{X$_{pre}$} ($X$ is the bigger dataset's name, e.g., \textbf{CNNDM$_{pre}$}) in our result tables. This is analogous to the continual training method proposed for multilingual transfer learning of machine translation \cite{kocmi2018trivial}. (2) \emph{Joint-training}, in which we upsample \textsc{EmailSum} data and mix it with another dataset, then use the combined data to finetune T5, similarly denoted as \textbf{X$_{joint}$}. This is analogous to the multilingual joint training method used in machine translation \cite{johnson2017google}.

\paragraph{Semi-supervised learning.} 
Since we only have 2.5K labeled email threads, another important technique to improve the performance is to utilize unlabeled data (i.e., email threads without labeled summaries). As introduced in Section~\ref{sec:data1}, in addition to the 3K email threads used for summary collection, we have 8,594 unlabeled email threads (5,116 from Avocado; 3,478 from W3C). We explore semi-supervised learning via the simple self-training technique \cite{scudder1965probability}. We use a trained model (a finetuned T5) to generate summaries for unlabeled threads, then mix the model-labeled and human-labeled data to finetune T5 again, referred as \textbf{SemiSup$_x$} ($x$ stands for the unlabeled data source we use, i.e., W3C, Avocado, or together).

\begin{figure}
    \centering
    \includegraphics[width=0.47\textwidth]{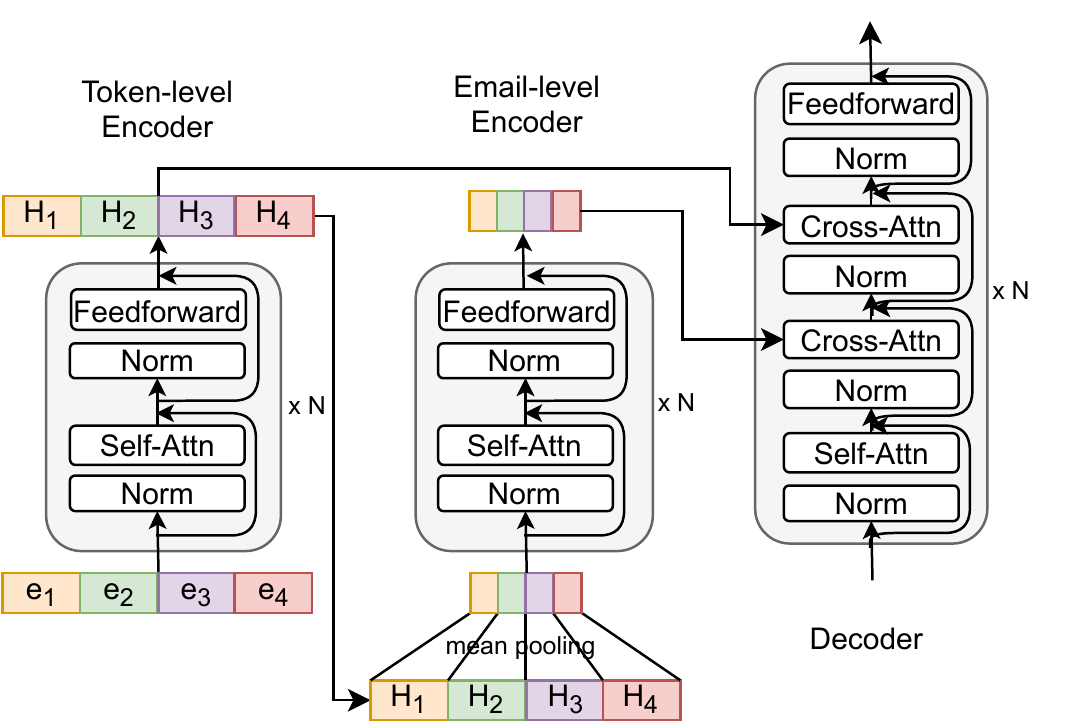}
    \vspace{-3pt}
    \caption{The architecture of our hierarchical T5.}
    \label{fig:ht5}
    \vspace{-10pt}
\end{figure}

\paragraph{Hierarchical T5.} Hierarchical summarization models have been shown to improve the performance of multi-document summarization task \cite{liu2019hierarchical}. Although an email thread can be treated as a single document due to the temporal dependency between consecutive emails, it also has a clear turn structure that encourages using of the hierarchical encoders. Recently, \citet{zhu2020hierarchical} proposed a hierarchical model (HMNet) for meeting summarization. Inspired by their work, we propose a hierarchical model that is similar to HMNet in structure but uses T5 as the backbone, therefore, it can take advantage of both the hierarchical structure and the pre-trained knowledge. As shown in Figure~\ref{fig:ht5}, this model contains two encoders: the \emph{token-level} encodes the whole email thread (e.g., $e_1, e_2, e_3, e_4$) while the \emph{email-level} receives mean-pooled email-level representations as input. The decoder has two cross attentions that attend to the outputs of the email-level and the token-level encoders respectively. Both token-level and email-level encoders are sharing the weights of the T5 encoder. We add a small number of new parameters by adding new cross attention between the decoder and the email-level encoder. 

%% file: experiments.tex
\begin{savenotes}
\begin{table*}
\begin{center}
\small
\begin{tabular}{lllllllllll}
\toprule 
Models & \multicolumn{5}{c}{\textsc{EmailSum}$_{short}$} & \multicolumn{5}{c}{\textsc{EmailSum}$_{long}$} \\
 \cmidrule(lr){2-6} \cmidrule(lr){7-11}
& R1 & R2 & RL & RLsum & BertS & R1 & R2 & RL & RLsum & BertS \\
 \midrule
 \emph{Oracle} &  \emph{39.04} &  \emph{12.47} &  \emph{30.17} &  \emph{35.61} & \emph{22.32} &  \emph{45.98} &  \emph{15.49} &  \emph{32.40} &  \emph{42.14} & \emph{26.31}  \\
Lead-1 & 23.35  & 5.57 & 18.22 & 19.61 & 12.25 & 19.75 & 4.84 & 14.24 & 16.88 & 6.87 \\
Lead-1-Email & 26.62 & 5.60 & 19.72 & 23.77 & 13.00 & 35.71 & 8.69 & 24.70 & 32.13 & 16.93\\
TextRank & 22.52 & 4.54 & 16.56 & 20.24 & 5.89 & 28.42 & 6.20 & 19.08 & 25.19 & 5.67\\
BertSumExt & 24.84 & 5.15 & 17.81 & 21.81 & 7.51 & 30.23 & 7.08 & 19.59 & 26.68 & 7.78\\
 \midrule
Fast Abs RL  & 31.15 & 6.59 & 22.73 & 29.03 & 6.49 & 39.35 & 10.58 & 27.01 & 36.51 & 10.03\\
T5$_{base}$ & 36.57 & 10.56 & 28.3 & 32.76 & 33.90 & 43.81 & 14.08 & 30.47 & 39.88 & 32.09 \\
\cmidrule(lr){1-1}
\ \ CNNDM$_{pre}$ & 35.43 & 10.75 & 27.49 & 32.15 & 33.61 & 44.15 & 14.20 & 30.84 & 40.21 & 32.53 \\
\ \ XSum$_{pre}$ & 36.14 & 10.26 & 28.66 & 33.47 & \bf 33.97 & 43.48 & 13.82 & 30.14 & 39.80 & 31.60 \\
\ \ SAMSum$_{pre}$ & 34.68 & 10.56 & 26.62 & 31.22 & 33.25 & 42.83 & 13.54 & 30.00 & 39.13 & 31.82  \\
\ \ CRD3$_{pre}$ &  36.05 & 10.04 & 27.21 & 32.06 & 33.52 & 43.60 & 13.93 & 30.49 & 39.97 & 31.53   \\
\cmidrule(lr){1-1}
\ \ CNNDM$_{joint}$ & 34.38 & 9.27 & 27.20 & 31.30 & 32.70 & 43.28 & 12.37 & 28.84 & 39.39 & 29.95 \\
\ \ XSum$_{joint}$ & 34.18 & 8.17 & 25.94 & 30.68 & 31.83 & 42.36 & 11.85 & 28.23 & 38.31 & 29.22 \\
\ \ SAMSum$_{joint}$ & 35.57 & 10.07 & 27.95 & 32.57 & 33.55 & 42.96 & 13.44 & 29.99 & 39.54 & 31.82 \\
\ \ CRD3$_{joint}$ & 34.66 & 8.81 & 26.95 & 31.59 & 33.29 & 42.81 & 12.96 & 29.35 & 39.33 & 32.14  \\
\cmidrule(lr){1-1}
\ \ SemiSup$_{w3c}$ & 35.43 & 10.64 & 28.59 & 32.31 & 33.61 & \bf  44.56$^{**}$ & 14.60$^{**}$ & \bf 31.38$^{**}$ & \bf 40.73$^{**}$ & \bf 32.81$^{**}$ \\
\ \ SemiSup$_{avocado}$ & 36.73 & 10.82 & 28.44 & 33.25 & 33.76 & 43.83 & \bf 14.61$^{**}$ & 31.21$^{**}$ & 40.52$^{*}$ &  32.71$^{**}$ \\
\ \ SemiSup$_{together}$ & \bf 36.98 & \bf 11.21$^{*}$ & \bf 28.76 & \bf 33.70$^{**}$ & 33.91 & 44.08 & 14.06 & 31.17$^{**}$ & 40.67$^{**}$ & 32.30 \\
\cmidrule(lr){1-1}
Hier. T5$_{base}$ & 36.17 & 10.37 & 28.44 & 33.34 & 33.39 & 44.50$^{*}$ & 14.53$^{*}$ & 30.89$^{*}$ & 40.22 & 32.30 \\
\toprule
\end{tabular}
\end{center}
\vspace{-12pt}
\caption{Summarization performance on the testing set of different models. We test the significance\footnote{The significance test is following the bootstrap test setup \cite{efron1994introduction} and sample for 100k times.} of the improvement over T5$_{base}$ ($*$: $p<0.05$, $**$: $p<0.01$). 
}
\label{tab:test}
\vspace{-12pt}
\end{table*}
\end{savenotes}

\section{Experiments}
\subsection{Evaluation Metrics}
\paragraph{ROUGE} \cite{lin2004rouge} is 
a commonly used automatic metric for summarization tasks. It has several variants: (1) ROUGE-1 (R1) measures the unigram overlap between the generated and reference summaries; (2) ROUGE-2 (R2) measures the bi-gram overlap; (2) ROUGE-L (RL) computes the longest common subsequence (LCS); (4) summary-level ROUGE-L (RLsum) computes LCS between each pair of reference and candidate sentences and returns the union-LCS. We use the \texttt{rouge\_score} package\footnote{\url{https://github.com/google-research/google-research/tree/master/rouge}} and report F1 scores. 

\paragraph{BERTScore} \cite{zhang2019bertscore} goes beyond n-gram overlap to provide contextualized semantic similarity.
Specifically, it uses BERT \cite{devlin-etal-2019-bert} (or RoBERTa \cite{liu2019roberta}) representations to ``softly'' align the words in candidate and reference summaries and then computes a ``soft'' uni-gram F1 score. We use the \texttt{bert\_score} package\footnote{\url{https://github.com/Tiiiger/bert_score}} and report rescaled numbers with a baseline.

\subsection{Results}
Table~\ref{tab:test} shows the evaluation results on the testing set of different models (the corresponding results on the development set can be found in Appendix Table~\ref{tab:dev}). It can be observed that the \emph{Oracle} extractive model sets up a high upper bound on all metrics except for BERTScore (BertS). 
Among non-oracle extractive methods, the \emph{Lead-1-Email} heuristic works best and even better than the deep extractive method, BertSumExt. The hybrid \emph{Fast Abs RL} model outperforms purely extractive methods but works worse than purely abstractive methods with large-scale pretraining (e.g., T5).

\begin{savenotes}
\begin{table}
\begin{center}
\small
\begin{tabular}{ccccc}
\toprule 
& \multicolumn{2}{c}{\textsc{EmailSum}$_{short}$} & \multicolumn{2}{c}{\textsc{EmailSum}$_{long}$} \\
 \cmidrule(lr){2-3} \cmidrule(lr){4-5}
& EO-R1$\downarrow$ & LE-R1$\downarrow$ & EO-R1$\downarrow$ & LE-R1$\downarrow$ \\
\midrule
Human & 39.0 & 26.62 & 46.0 & 35.71 \\
T5$_{base}$ & 50.27 & 36.88 & 55.43 & 43.65 \\
R1-best & 52.50 & 39.22 & 60.04 & 49.14  \\ 
\toprule
\end{tabular}
\end{center}
\vspace{-12pt}
\caption{The extractive Oracle (EO) and Lead-1-Email (LE) models' ROUGE-1 by taking human summary, base or best model generated summary as the ground-truth. The lower the scores are, the more abstractive the summaries are ($\downarrow$).}
\label{tab:ext}
\vspace{-13pt}
\end{table}
\end{savenotes}

Taking the email thread as one single document and finetuning T5 (i.e., T5$_{base}$ in Table~\ref{tab:test}) sets up a strong baseline. Upon this baseline model, we test the transfer learning from four different summarization datasets (CNN/DM, XSum, SAMSum, and CRD3). However, as shown in Table~\ref{tab:test}, transfer learning barely improves over baseline, and transferring by \emph{pre-finetuning} always works better than \emph{joint-training}. Since our \textsc{EmailSum} has a quite different domain as existing news or dialogue datasets, we conjecture that it is hard to transfer knowledge between them or better transferring techniques need to be applied.
Similarly, we test the semi-supervised learning with unlabelled data from W3C, Avocado, and both of them (together). This method can mostly (or significantly in some cases) outperform the baseline's performance for both \textsc{EmailSum}$_{short}$ and \textsc{EmailSum}$_{long}$. Lastly, the hierarchical T5$_{base}$ model only marginally outperforms the non-hierarchical baseline
for \textsc{EmailSum}$_{long}$ task. It is notable that overall \textsc{EmailSum}$_{long}$ has higher ROUGE scores but lower BERTScore than \textsc{EmailSum}$_{short}$.

\begin{figure}
    \centering
    \includegraphics[width=0.40\textwidth]{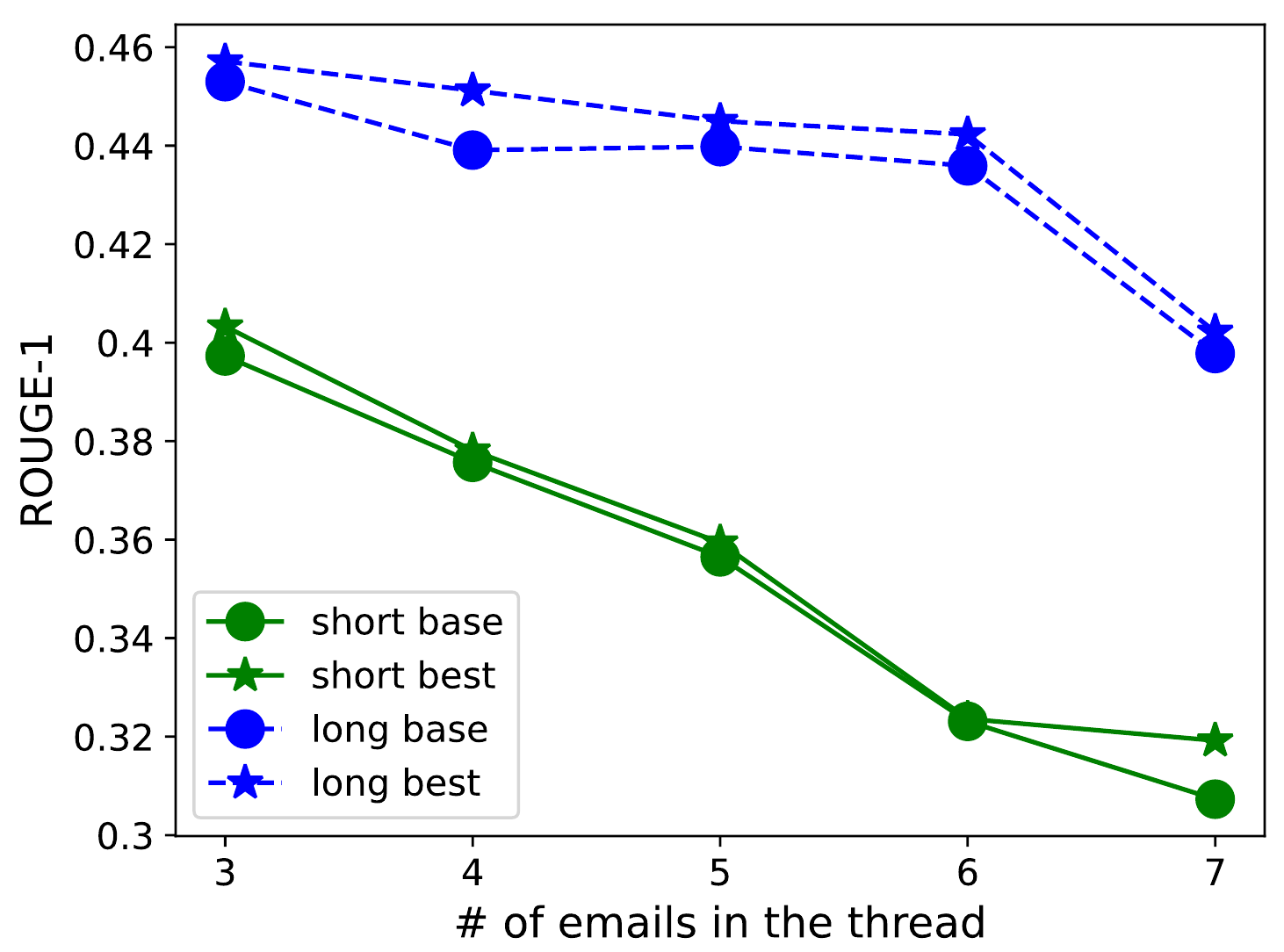}
    \vspace{-5pt}
    \caption{The impact of the number of emails in the thread on summarization performance (ROUGE-1). The results are on the testing set. short/long denotes \textsc{EmailSum}$_{short}$/\textsc{EmailSum}$_{long}$; base/best denotes the baseline/best model.}
    \label{fig:thread_length}
    \vspace{-15pt}
\end{figure}

\begin{savenotes}
\begin{table*}
\begin{center}
\small
\begin{tabular}{ccccccc}
\toprule 
& \multicolumn{3}{c}{\textsc{EmailSum}$_{short}$} & \multicolumn{3}{c}{\textsc{EmailSum}$_{long}$} \\
 \cmidrule(lr){2-4} \cmidrule(lr){5-7}
& \multicolumn{3}{c}{SemiSup$_{together}$ vs T5$_{base}$} &  \multicolumn{3}{c}{SemiSup$_{w3c}$ vs T5$_{base}$} \\
 \midrule
& Win & Lose & Tie & Win & Lose & Tie \\
Salience & 109 & 133 & 55 & 109 & 130 & 50\\
Faithfulness & 116 & 123 & 58 & 126 & 122 & 41 \\
Overall quality & 120 & 138 & 39 & 125 & 140 & 24 \\
\toprule
\end{tabular}
\end{center}
\vspace{-12pt}
\caption{Pairwise comparison between summaries generated by the best ROUGE-1 models and T5$_{base}$.}
\label{tab:human}
\vspace{-8pt}
\end{table*}
\end{savenotes}

Since we focus on generating abstractive summaries for email threads and the human-written summaries are fairly abstractive (as shown in Table~\ref{tab:stats}), we further investigate the abstractiveness of model-generated summaries. We take summaries generated by the baseline (T5$_{base}$) and the best ROUGE-1 models (SemiSup$_{together}$ for \textsc{EmailSum}$_{short}$, SemiSup$_{w3c}$ for \textsc{EmailSum}$_{long}$) as the pseudo ground-truth, respectively. Then, we evaluate the ROUGE-1 of extractive \emph{Oracle} and \emph{Lead-1-Email} models; higher scores means more extractive summaries. As shown in Table~\ref{tab:ext}, compared to humans, models generate much more extractive summaries. Moreover, the semi-supervised models (R1-best) are even more extractive than the baseline, which is probably because the self-training procedure amplifies the extraction tendency. Lastly, for both base and best models as well as for both short and long summaries, the model performance (ROUGE-1) decreases as the number of emails in the thread increases (shown in Figure~\ref{fig:thread_length}).

%% file: human_eval.tex
\section{Human Evaluation}
\label{sec:human_eval}
\subsection{Human Rating Collection}
To better understand where the model still falls short and investigate if the automatic metrics correlate well with human judgments, we conduct a human evaluation on Amazon Mechanical Turk. Initially, by manually checking the quality of model-generated summaries, we find that models can mostly generate grammatical, relevant, and fluent summaries; however, they often fail to be salient and faithful, i.e., models tend to be over-detailed or do not stay true to the source thread. Therefore, we ask human annotators to rate the ``salience'' and ``faithfulness'' of model-generated summaries. We choose the best ROUGE-1 models, SemiSup$_{together}$ for \textsc{EmailSum}$_{short}$, SemiSup$_{w3c}$ for \textsc{EmailSum}$_{long}$, to evaluate, then we sample 100 examples, and collect 3 responses for each example. Human judges are asked to rate on a 5-point Likert scale for salience and faithfulness respectively and annotate which summary sentences are not salient or unfaithful. We explain the meaning of ``salience'' and ``faithfulness'' to annotators and instruct them how to rate from 1 to 5. Meanwhile, to verify the improvement obtained by best R1 models over T5$_{base}$, we ask them to compare the summaries generated by these models and those from T5$_{base}$, and judge which one is more salient, more faithful, and has overall higher quality. More collection details can be found in the Appendix~\ref{appendix:human}.

We check the average inter-rater agreement (Krippendorff’s alpha \cite{krippendorff2011computing}) of ``salience'' and ``faithfulness'' ratings. It is around 0.09 to 0.23, i.e., slight to fair agreement \cite{fleiss1973equivalence}. However, when we convert the ratings to 3-point by taking \{3\}, \{4 and 5\}, \{1 and 2\} as 3 classes, the agreement increases to 0.36 to 0.63, i.e., fair to substantial agreement. This indicates that humans' subjectivity affects the ratings and people have a hard time distinguishing `bad' from `very bad' as well as `good' from `very good'. Meanwhile, the ratings for short summaries are always less agreed across raters (0.36-0.38) than that for long summaries (0.58-0.63). This indicates that there might be multiple different ways of summarizing an email thread into a short summary. The agreement of pairwise comparison is around 0.20 to 0.24 (fair agreement), which is because the baseline and the best models have non-distinguishable performance (shown in Table~\ref{tab:human}). Finally, we take the 3-rater average as the final human rating for each example. 

In addition, we evaluate the correlations (\emph{Pearson Correlation} \cite{benesty2009pearson}) among different human ratings. The correlation between salience and faithfulness ratings is 0.36/0.45 for short/long summarization. And the correlations among salience, faithfulness, and overall quality pairwise preferences are around 0.53 to 0.79. Overall, moderate to large \cite{cohen2013statistical} correlations are observed. 

\begin{table*}
\begin{center}
\small
\begin{tabular}{p{\textwidth}}
\toprule 
\textbf{Fail to understand the sender's intent.} \\
\midrule
\emph{Thread}: Subject: minutes of meeting: 3.5 plan $\vert\vert\vert$ Om: 1. Nihar mentioned that we spent about 3 weeks in redefining the language, which was not originally planned. This is the major reason for moving the code freeze date from 8/24 to 9/21. 2. For phase-I code drop to QA on 8/28 The confidence in date is : 90\% The confidence in statbility of build is : 80\%  3. ... $\vert\vert\vert$ Sharon: Hi Om - We also need to lock down the date for: 1 - service pack merge 2 - bug fix freeze and, Javascript library testing (Offline) resource thanks, sharon $\vert\vert\vert$ Rajeev: Thanks for the meeting minutes. Nihar, Sharon can you list the Risks to the phase 1 \& Phase II schedules and what we are doing to manage the risk. Rajeev \\
\emph{Generated Summary}: \colorbox{red!15}{Om tells Nihar that he spent 3 weeks redefining the language.} 
Sharon tells Om that she needs to lock down the date for 1 - service pack merge 2 - bug fix freeze and Javascript library testing. (\textbf{salience=4, faithfulness=3.3})\\
\emph{Ground-truth}: \colorbox{green!15}{Om gives everyone minutes for a meeting.} Sharon updates Om on some other plans and Rajeev asks Nihar/Sharon for some technical details. \\
\midrule
\midrule
\textbf{Fail to identify the roles of the sender and receiver.} \\
\midrule
\emph{Thread}: Subject: latest 4.0 ga palladium install for biogen $\vert\vert\vert$ Nilesh: PATH/patchinstaller I tested this with build version 377 and it works fine. $\vert\vert\vert$ Diana: This one looks good. I have verified that the 2 fixes in 382 are in the patch installer. Just to clarify, this is really a 382 patch installer that falls under the 377 directory? ... $\vert\vert\vert$ Nilesh: Wilhan, I have deleted build 382 as there was no space to create patch installer. (as we discussed in the lab) And as we specified the build version to be 377 when creating the patch installer I thought we will need to put it under build 377 and use the jar files for that. Can you please clarify this. ...\\
\emph{Generated Summary}: \colorbox{red!15}{Nilesh tells Diana that the 2 fixes in 382 are in the patch installer.} Nileshe also asks Wilhan to clarify the definition of the build. (\textbf{salience=3.3, faithfulness=3.3}) \\
\emph{Ground-truth}: \colorbox{green!15}{Nilesh says he tested something with a build.} Diana thinks it looks good after verifying it but asks some questions. Nilesh updates Wilhan and has some questions. \\
\toprule
\end{tabular}
\end{center}
\vspace{-12pt}
\caption{Error analysis examples. Emails are separated by `$\vert\vert\vert$' and some content is omitted by `...'. (\textbf{salience=xx, faithfulness=xx}) gives the average human rating for that summary.}
\vspace{-5pt}
\label{tab:error-example}
\end{table*}

\subsection{Generated Summary's Quality Analysis}
Surprisingly, human evaluators are mostly satisfied with the salience and faithfulness of model-generated summaries, ratings are around 4 out of 5.
On average, humans rate 3.89 and 4.04 for the salience and faithfulness of SemiSup$_{together}$ generated short summaries, respectively; and they rate 4.22 and 4.29 for the salience and faithfulness of SemiSup$_{w3c}$ generated long summaries, respectively. Examples with low or high ratings are shown in Table~\ref{tab:error-example} or
Appendix Table~\ref{tab:high-example}. Humans rate higher for model-generated long summaries, which is correlated to the trend of ROUGE, and they are more satisfied with faithfulness than salience.

Table~\ref{tab:human} presents the human pairwise comparison between the best ROUGE-1 models and T5$_{base}$. Except for the faithfulness of \textsc{EmailSum}$_{long}$, the best ROUGE-1 models mostly lose to the baseline (though the loss and win are mostly marginal). Together with Table~\ref{tab:ext}, we conjecture that the improvement obtained by semi-supervised learning exploits n-gram matching accuracy by making the summary more extractive, while humans prefer more abstractive summaries.

Lastly, we analyze the non-salient and unfaithful sentences labeled by the human evaluators. We find that two errors are frequently made by the summarization model: \textbf{(1) Failing to understand the sender's intent.} Usually, when we send an email, there is a high-level intention behind the detailed content we write, e.g., start up a discussion, bring up a concern, broadcast a decision, etc. However, models are oftentimes unable to capture the intention and thus overly focus on details. As shown in the first example of Table~\ref{tab:error-example}, \textit{Om} intends to summarize the important points from a meeting, while the model only picks the first piece of detail in that email as the summary. This problem is also related to the over-extractive issue (shown in Table~\ref{tab:ext}). The model tends to extract details from the source thread and the extraction is biased to the first sentence of each email. \textbf{(2) Failing to identify the roles of the sender and receiver.} An email thread is a special type of conversation with multiple speakers involved. One important task for the model is to identify the roles of different speakers and their relations, i.e., who does what to whom. As shown in the second example of Table~\ref{tab:error-example}, the model wrongly takes ``2 fixes in 382 are in the patch installer'' as information provided by \textit{Nilesh}, whereas it is supposed to be by \textit{Diana}. The same issue can also be observed in the first example: \textit{Om} is just summarizing what \textit{Nihar} said instead of telling \textit{Nihar}. This is considered as a type of unfaithfulness, which has been widely identified as a common issue of abstractive summarization models \cite{wang2020asking, durmus2020feqa, maynez2020faithfulness}. 

\begin{figure}
    \centering
    \includegraphics[width=0.48\textwidth]{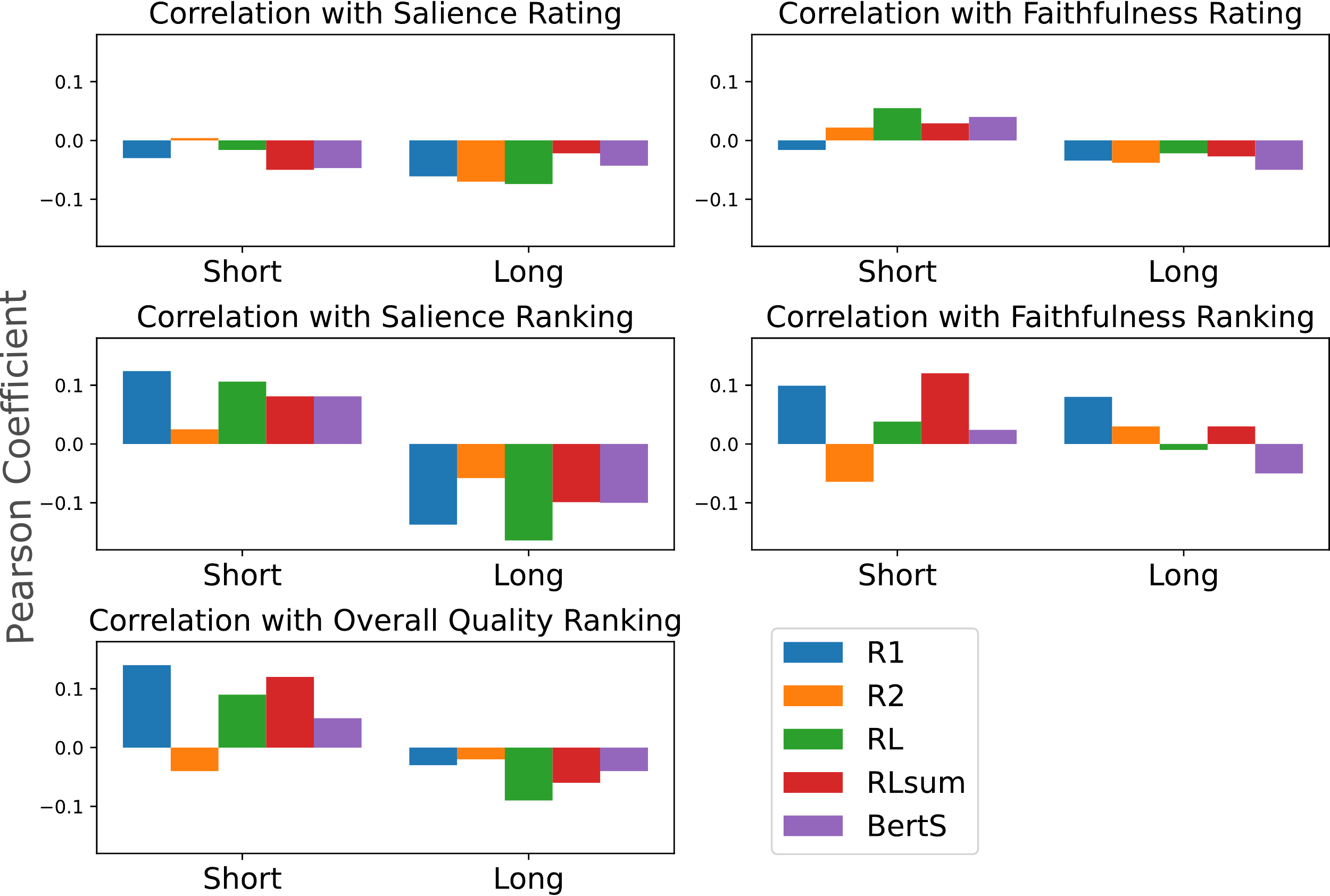}
    \vspace{-15pt}
    \caption{Correlation between automatic metrics and human judgements. Short and Long refer to \textsc{EmailSum}$_{short}$ and \textsc{EmailSum}$_{long}$ tasks, respectively. }
    \label{fig:correlation}
    \vspace{-10pt}
\end{figure}

\subsection{Correlation with Human Judgement}
ROUGE \cite{lin2004rouge} measures n-gram overlap and BERTScore \cite{zhang2019bertscore} is essentially based on ``soft'' uni-gram matching. However, according to our analysis presented above, the email thread summarization models mainly fail to be abstractive, salient, and faithful, which are hard to be evaluated by n-gram overlap. Furthermore, as pointed out by \citet{bhandari-etal-2020-evaluating}, different datasets usually require different evaluation metrics. Therefore, here, we study the correlation between automatic metrics and human judgments.

Specifically, we evaluate the \emph{Pearson Correlation} between human ratings and automatic metric scores on the 100 examples used in the human evaluation. Besides, as described above, we conduct a pairwise model comparison between the best ROUGE-1 models and T5$_{base}$ for ``salience'', ``faithfulness'', and ``overall quality''. We convert them to a pairwise ranking score, i.e., -1 if T5$_{base}$ is better; 1 if T5$_{base}$ is worse; 0 if two models are non-distinguishable.
In the same way, we convert different metric scores to ranking scores. Then, we also evaluate the \emph{Pearson Correlation} between human and metric ranking scores. Figure~\ref{fig:correlation} illustrates the results. Overall, the correlations are fairly poor. The best correlation is between ROUGE-1 and human overall quality ranking for short summary generation (coefficient=0.14, p=0.16). There is little or negative correlation between metrics and human judgment for the long summary generation. Therefore, we emphasize the importance of human evaluation and better automatic proxies need to be proposed in the future. 

%% file: appendix.tex
\appendix
\section*{Appendix}
\section{Summary Collection}
\label{appedix:summay_annotation}
\begin{figure*}
    \centering
    \includegraphics[width=0.88\textwidth]{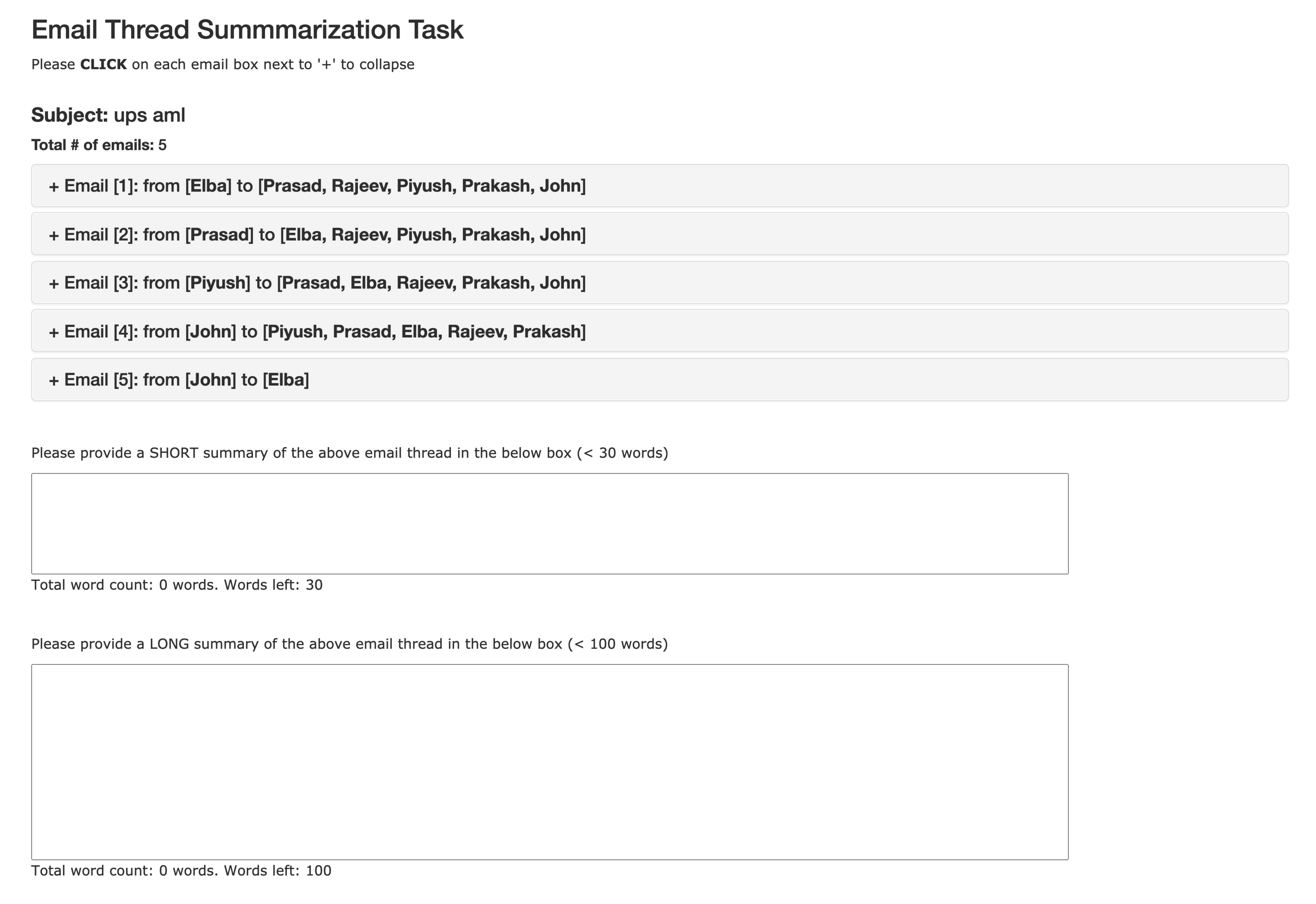}
    \vspace{-3pt}
    \caption{A part of the Amazon Mechanical Turk webpage used for collecting summaries.}
    \label{fig:summary_collection}
    \vspace{-10pt}
\end{figure*}

Figure~\ref{fig:summary_collection} illustrates the questions we asked human annotators on Amazon Mechanical Turk during summary collection. 
Before these questions, here are some important instructions we listed on the webpage: (1) Long summary MUST be longer than short summary; (2) Summary length can be dynamically decided based on the content of the thread; (3) Short summary should be a \emph{concise and abstractive} description of what the thread is mainly talking about; (4) Long summary can be a narrative of what happens. But do NOT simply summarize each email separately. The summary should be \emph{coherent}; (5) It is NOT necessary to summarize every email in the long summary, i.e., it is OK to skip unimportant ones and merge similar ones if needed; (6) You are \emph{encouraged} to include important sender and/or receiver names in long summary; (7) You are \emph{disencouraged} to copy a lot from emails for both short and long summaries; You are supposed to write in your own words as much as you can; (8) You may find some content are technical. We do NOT expect any background knowledge. Just focus on the major concerns, decisions, and consensus. (9) In the thread, emails are ordered by time. However, one email does NOT necessarily reply to the previous one. It can reply to an earlier email OR forward to new receivers. In other words, the structure is NOT always continuous, so please be careful when you read.

\section{Fast Abs RL}
\label{appedix:fastabs}
The original Fast Abs RL method \cite{chen2018fast} uses ROUGE-L$_{recall}$ to align extracted source sentences and target summary sentences. In our case, we extract emails and align them with summary sentences. Since the emails and summary sentences usually follow the same temporal order, we enhance the alignment procedure by the Neeleman-Wunsch algorithm \cite{needleman1970general, rameshkumar2020storytelling} to imposing strict order constraints, e.g., there should not be ``email$_i$ is aligned to sentence$_j$ while email$_{i+1}$ is aligned to sentence$_{j-1}$'' cases. Meanwhile, we modify it to allow one email to be aligned with multiple summary sentences but avoid one summary sentence aligning with multiple emails. Specifically, we first obtain the similarity matrix $M$ of size $n_e \times n_s$ between each email and summary sentence by ROUGE-L$_{recall}$ ($n_e$ is the number of emails, $n_s$ is the number of summary sentences); then the alignment score matrix $H$ of size $(n_e + 1) \times (n_s + 1)$ is initialized as all-zero then computed as follows for $1 \leq x \leq n_e$, $1 \leq y \leq n_s$:
\begin{align*}
    H_{x,y} = \max \left\{
                \begin{array}{ll}
                  H_{x-1, y-1} + M_{x-1, y-1}\\
                  H_{x, y-1} + M_{x-1, y-1}\\
                  H_{x-1, y}
                \end{array}
              \right.
\end{align*}
Then we traceback from $H_{n_e,n_s}$ to $H_{0,0}$ to obtain the final alignment. As shown in Table~\ref{tab:dev}, the ``Fast Abs RL (default)'' model refers to this method with the default setting which works mostly worse than our enhanced Fast Abs RL. 

\section{Experimental Details \& Additional Results}
\label{appedix:exp}
We implement the TextRank \cite{mihalcea2004textrank} model via the \texttt{summa} python package\footnote{\url{https://github.com/summanlp/textrank}} and set the summarization ratio as the average $\frac{summary\ length}{thread\ length}$ ratio in the training set, which is 0.22 for short summary and 0.38 for long summary. We test Fast Abs RL \cite{chen2018fast} via the author's open-source code.\footnote{\url{https://github.com/ChenRocks/fast_abs_rl}} Most of our models are built on T5 \cite{raffel2020exploring} and we use the base version that has 220 million parameters. Our hierarchical T5 shares the same T5 encoder parameters between the token-level and email-level encoders. The only new parameters added are from the first cross attention between decoder and email-level encoder. We use \texttt{Transformers} \cite{wolf-etal-2020-transformers}\footnote{\url{https://github.com/huggingface/transformers}} to run all the T5 based models. We run experiments on a single Tesla V100 GPU. We set the max input sequence length as 512 tokens and max output length as 56 tokens during training (200 tokens during evaluation). The total batch size (with gradient accumulation) is 128. The learning rate is 5e-4, except for training the T5$_{base}$ from scratch, we use 1e-4 instead. Since our training set only contains 1.8K examples,  it only takes 2-4 minutes per epoch. We train models for 70 epochs. 

Our model selection is based on each of the five evaluation metrics, ROUGE-1/ROUGE-2/ROUGE-L/summary-level ROUGE-L/BERTScore. We select the best checkpoints for each of the five metrics on our development set, then test those checkpoints on the testing set to report the final numbers for each metric. Table~\ref{tab:dev}
shows all the results on our development set. Table~\ref{tab:high-example} shows two examples that have high-rating model-generated summaries.

\begin{savenotes}
\begin{table*}
\begin{center}
\small
\begin{tabular}{lllllllllll}
\toprule 
Models & \multicolumn{5}{c}{\textsc{EmailSum}$_{short}$} & \multicolumn{5}{c}{\textsc{EmailSum}$_{long}$} \\
 \cmidrule(lr){2-6} \cmidrule(lr){7-11}
& R1 & R2 & RL & RLsum & BertS & R1 & R2 & RL & RLsum & BertS \\
 \midrule
 \emph{Oracle} &  \emph{39.8} &  \emph{13.05} &  \emph{31.17} &  \emph{36.15} & \emph{28.50} &  \emph{46.74} &  \emph{17.13} &  \emph{33.92} &  \emph{43.1} & \emph{28.38}  \\
Lead-1 & 25.63  & 6.56 & 19.97 & 21.51 & 13.55 & 20.72 & 5.87 & 15.23 & 18.01 & 8.09 \\
Lead-1-Email & 26.37 & 5.88 & 19.68 & 23.61 & 12.98 & 36.65 & 10.44 & 26.00 & 33.27 & 18.11 \\
TextRank & 21.91 & 4.20 & 16.12 & 19.57 & 6.56 & 29.00 & 7.15 & 20.00 & 25.92 & 10.44 \\
BertSumExt & 25.76 & 6.02 & 18.74 & 22.59 & 8.34 & 30.90 & 8.29 & 20.91 & 27.55 & 8.92 \\
 \midrule
Fast Abs RL (default)  & 29.67 & 6.08 & 22.68 & 27.66 & 6.92 & 39.43 & 11.08 & 25.78 & 36.81 & 7.14 \\
Fast Abs RL  & 31.56 & 6.52 & 23.01 & 29.51 & 5.59 & 39.24 & 11.25 & 27.77 & 36.72 & 9.63 \\
T5$_{base}$ (from scratch) & 19.71 & 1.95 & 14.88 & 16.75 & 22.52 & 24.51 & 3.72 & 15.72 & 21.91 & 9.70 \\
T5$_{base}$ & 36.78 & 11.93 & 29.50 & 33.58 & 34.92 & 44.94 & 15.94 & 32.33 & 41.22 & 33.67 \\
\cmidrule(lr){1-1}
\ CNNDM$_{pre}$ & 37.00 & 11.26 & 28.97 & 33.49 & 35.09 & 44.83 & 15.88 & 32.02 & 41.25 & 33.89 \\
\ XSum$_{pre}$ & 36.63 & 11.43 & 29.43 & 33.75 & 35.29 & 44.55 & 15.29 & 31.50 & 40.87 & 33.47 \\
\ SAMSum$_{pre}$ & 36.72 & 11.1 & 28.73 & 33.21 & 35.82 & 44.31 & 15.36 & 31.45 & 40.63 & 33.60  \\
\ CRD3$_{pre}$ &  36.84 & 11.57 & 29.19 & 33.38 & 35.37 & 44.57 & 15.73 & 31.87 & 40.91 & 33.47   \\
\cmidrule(lr){1-1}
\ CNNDM$_{joint}$ & 35.89 & 10.41 & 28.02 & 32.41 & 34.02 & 43.92 & 14.48 & 30.54 & 39.99 & 31.67 \\
\ XSum$_{joint}$ & 35.07 & 9.26 & 27.18 & 31.53 & 34.27 & 43.36 & 13.35 & 29.44 & 39.45 & 30.97 \\
\ SAMSum$_{joint}$ & 36.59 & 11.20 & 29.20 & 33.49 & 35.44 & 44.38 & 15.23 & 31.68 & 40.69 & 33.65 \\
\ CRD3$_{joint}$ & 36.24 & 10.43 & 28.55 & 32.72 & 35.52 & 44.25 & 14.87 & 31.24 & 40.38 & 33.57  \\
\cmidrule(lr){1-1}
\ SemiSup$_{w3c}$ & 37.03 & 11.92 & 29.30 & 33.78 & 35.60 & 45.03 & 16.09 & 32.50 & 41.52 & 33.95 \\
\ SemiSup$_{avocado}$ & 37.78 & 12.56 & 30.09 & 34.50 & 34.88 & 45.49 & 16.21 & 32.97 & 41.82 & 34.42 \\
\ SemiSup$_{together}$ & 37.43 & 12.26 & 29.84 & 34.32 & 35.08 & 45.73 & 16.27 & 32.65 & 41.91 & 34.09 \\
\cmidrule(lr){1-1}
Hier. T5$_{base}$ & 36.67 & 11.79 & 29.13 & 33.58 & 35.71 & 45.26 & 16.13 & 32.62 & 41.55 & 33.99 \\
\toprule
\end{tabular}
\end{center}
\vspace{-12pt}
\caption{Summarization performance of different models on the development set.}
\label{tab:dev}
\vspace{-12pt}
\end{table*}
\end{savenotes}

\begin{table*}
\begin{center}
\small
\begin{tabular}{p{\textwidth}}
\toprule 
\textbf{Examples of high-quality summaries generated by the model.} \\
\midrule
\emph{Thread}: Subject: faa demos $\vert\vert\vert$ Dan: PM Team, Attached are some general ideas and issues around developing new demos for our new target markets. Please review and provide feedback. Also, please provide links where we can learn more about various FAA applications. Thanx, Dan. $\vert\vert\vert$ Dan, Thanks for putting the high level descriptions together. My questions are: *Is it practical to do an EAI demo given the inherent complexity of application integration? ... *Should we delay looking at Outlook for now?... *What do you think that timelines are developing these demos? ... Alex $\vert\vert\vert$ Alex, Thanks for the feedback, please see my comments below:\\
\emph{Generated Short Summary}: Dan asks the PM team to review and provide feedback on FFA demos. Alex responds with questions. Dan thanks Alex and gives his feedback. (\textbf{salience=4.3, faithfulness=4.7}) \\
\emph{Ground-truth}: Dan talks about general ideas about demos to his PM team. Alex provides some feedback and asks questions. Dan thanks Alex for the feedback and adds comments. \\
\midrule
\emph{Thread}: Subject: sun performance report $\vert\vert\vert$ Mahesh: Hi, I am attaching the draft of the performance/sizing report for EMAS on Sun. Please send me your comments. I am also attaching a list of features that would be good to have. Thanks, Mahesh $\vert\vert\vert$ Amitabh: do we have a side-by-side comparison of solaris, hp-ux, and nt? also, a price-performance comparison might also be useful  $\vert\vert\vert$ Rajeev: Dan, Please consider Amitabh's suggestions for the sizing requirement document that you are prepaing... $\vert\vert\vert$ Mahesh: we do not have comparison stats. It would be good to have them. $\vert\vert\vert$ Dan: Good points, we should have side-by-side comparisons and also price/performance...\\
\emph{Generated Long Summary}: Mahesh is attaching a draft of the performance/sizing report for EMAS on Sun and asking for comments.
Amitabh asks if there is a side-by-side comparison of solaris, hp-ux, and nt.
Rajeev asks Dan to consider Amibh's suggestions for the sizing requirement document.
Mahesesh says there are no comparison stats, but it would be good to have them.
Dan says there should be side- by-side comparies and also price/performance. (\textbf{salience=4.3, faithfulness=5})\\
\emph{Ground-truth}: Mahesh shows everyone a performance report for a future meeting and attaches his feedback. Amitabh gives feedback which Rajeev asks Dan to consider in a different task. Mahesh and Dan make suggestions about comparisons. \\
\bottomrule
\end{tabular}
\end{center}
\vspace{-12pt}
\caption{Examples of high-quality summaries generated by model. Emails are separated by `$\vert\vert\vert$' and some content are omit by `...'. (\textbf{salience=xx, faithfulness=xx}) gives the average human rating for that summary.}
\vspace{-12pt}
\label{tab:high-example}
\end{table*}

\section{Human Evaluation}
\label{appendix:human}
\begin{figure*}
    \centering
    \includegraphics[width=0.95\textwidth]{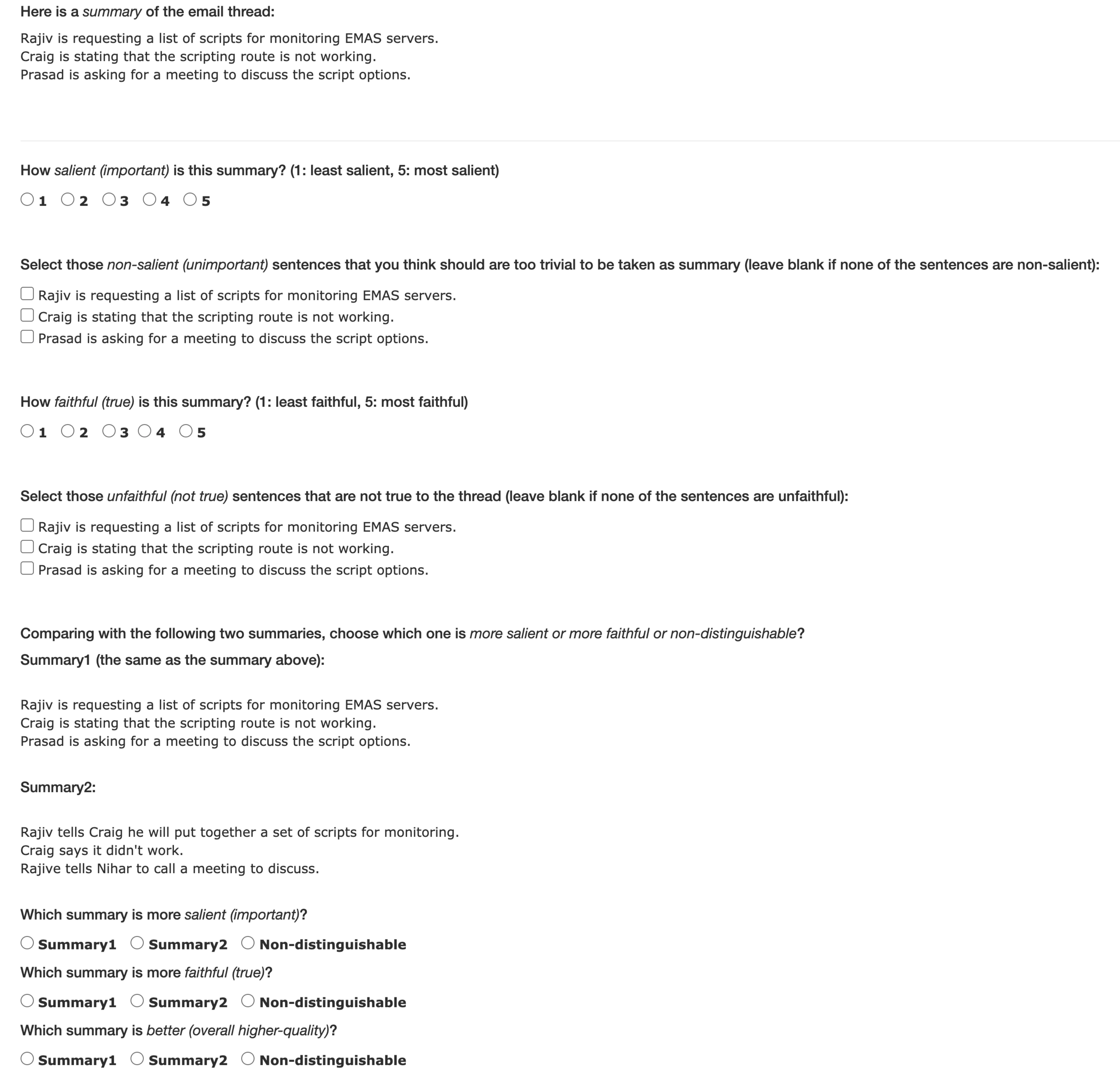}
    \vspace{-10pt}
    \caption{A part of the Amazon Mechanical Turk webpage used for human evaluation.}
    \label{fig:human_eval}
    \vspace{-12pt}
\end{figure*}

Figure~\ref{fig:human_eval} shows the questions we asked to human judges to evaluate the quality of model-generated summaries. Before these questions, we instruct annotators how to rate on a 5-point Likert scale for ``salience'' and ``faithfulness'': (1) Rate \emph{salience} from 1 to 5: 1 is the worst, none of the points in the summary is important enough to be summarized; 5 is the best, all of the points mentioned in the summary are important and worth to be summarized; (2) Rate \emph{faithfulness} from 1 to 5: 1 is the worst, all of the sentences in the summary are either wrong or not existing in the email thread; 5 is the best, all of the points mentioned in the summary are true to the thread. Plus, we also prompt examples of ``non-salient'' and ``unfaithful'' summaries on the webpage. We pay annotators \$0.60 per HIT.

